%% file: template.tex
\documentclass[journal]{vgtc}                     % final (journal style)
\vgtccategory{Research}

\vgtcpapertype{please specify}

\title{ASAP: Interpretable Analysis and Summarization of\\ AI-generated Image Patterns at Scale}

\author{Jinbin Huang 
\and Chen Chen
\and Aditi Mishra
\and Bum Chul Kwon
\and Zhicheng Liu
\and Chris Bryan}

\authorfooter{
\item
 {Jinbin Huang, Aditi Mishra and Chris Bryan} are with Arizona State University. E-mail: \{jhuan196, amishr45, cbryan16\}@asu.edu
\item
 Bum Chul Kwon is with IBM research. E-mail: bumchul.kwon@us.ibm.com.
\item
 {Chen Chen and Zhicheng Liu} are with University of Maryland. E-mail: \{cchen24,leozcliu\}@umd.edu.
}

\abstract{%
Generative image models have recently emerged as a promising technology that can produce realistic-looking images. Despite the potential benefits, there are growing concerns about its potential for misuse, particularly in generating deceptive images that could raise significant ethical, legal, and societal issues. Consequently, there is a growing recognition that we need to empower individual users to effectively discern and comprehend patterns of AI-generated images. To help achieve this goal, we have developed \sys, an interactive visualization system that automatically extracts distinct patterns of AI-generated images and allows users to interactively explore them using various views.
To uncover fake patterns, \sys introduces a novel image encoder, adapted and modified from CLIP, which transforms images into compact ``distilled'' representations, enriched with information for differentiating authentic and fake images. 
These representations are then used to generate gradients that propagate back to the attention maps of CLIP's transformer block. This process quantifies the relative importance of each pixel to image authenticity or fakeness, thereby exposing key deceptive patterns. \sys then enables the \textit{at scale} interactive analysis of these patterns through multiple, coordinated visualizations. This includes a representation overview with innovative cell glyphs to aid in the exploration and qualitative evaluation of fake patterns across a vast array of images, as well as a pattern view that not only displays authenticity-indicating patterns in images but also quantifies their impact. \sys supports the analysis of cutting-edge generative models with the latest architectures, including GAN-based models like proGAN and diffusion models like the latent diffusion model. We demonstrate the usefulness of \sys through two usage scenarios using several fake image detection benchmark datasets, which reveal its ability to identify and understand hidden patterns in AI-generated images, especially in detecting fake human faces produced by diffusion-based deepfake techniques.
}


\teaser{
  \centering
  \includegraphics[width=0.9\linewidth]{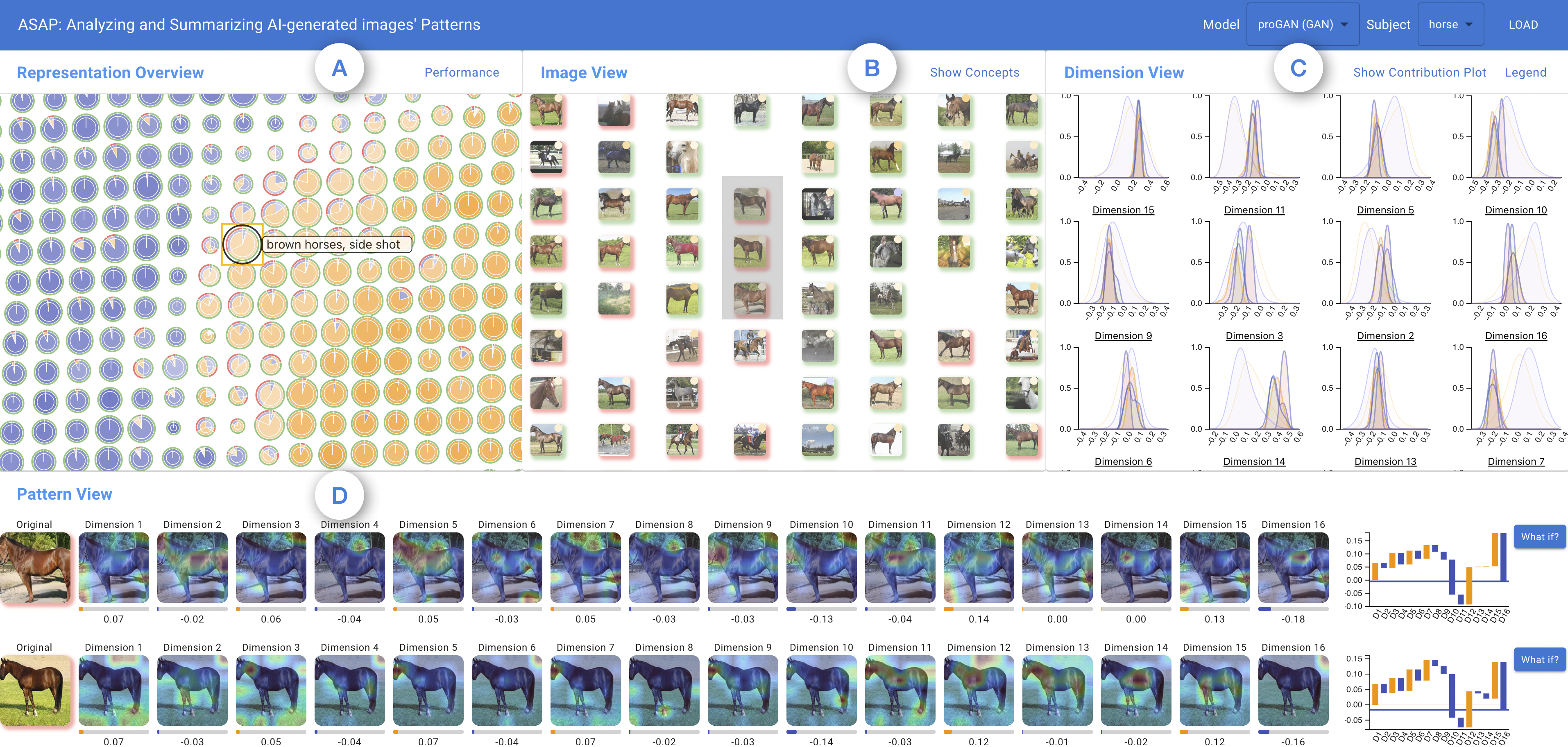}
  \caption{%
  	With \sys, users can efficiently analyze and identify deceptive patterns in AI-generated images that can fool trained classifiers. In this example, the user uses \sys to identify and investigate deceptive ``brown horses, side shot'' images generated by the proGAN model, which have successfully misled a trained fake image classifier. \sys provides an interactive interface with four main components: \textbf{(A) Representation Overview} utilizes tSNE to project image representations into a 2D space, grouping them into distinct, non-overlapping cells. These representations are derived from an image encoder trained to discern image authenticity. This view employs a novel glyph to summarize classification performances for images within each cell. Users can select each glyph to further inspect the patterns of the images within the corresponding cell using other views. \textbf{(B) Image View} offers a closer look at images within a selected cell, highlighting fake images that were classified as authentic. \textbf{(C) Dimension view} visualizes how images within a selected cell are distributed across the dimensions of the embeddings extracted from CLIP. This view helps users to correlate specific visual patterns with their quantitative dimensional distributions, enhancing the understanding of what makes certain images deceptive. \textbf{(D) Pattern View} facilitates a detailed analysis of individual images, allowing users to uncover the most misleading pixel groups both qualitatively and quantitatively. This pixel importance from the saliency map can reveal distinct patterns of AI-generated images.}
  \label{fig:teaser}
}

\graphicspath{{figs/}{figures/}{pictures/}{images/}{./}} % where to search for the images

\usepackage{tabu}                      % only used for the table example
\usepackage{booktabs}                  % only used for the table example
\usepackage{lipsum}                    % used to generate placeholder text
\usepackage{mwe}                       % used to generate placeholder figures
\usepackage{enumitem}
\usepackage{soul}
\usepackage{amsmath}
\usepackage{amssymb}
\usepackage{xspace}
\usepackage[svgnames, dvipsnames]{xcolor}

\definecolor{trueblue}{HTML}{0073cf}

\newcommand{\sys}{\textcolor{black}{\mbox{\textsc{Asap}}}\xspace}

\newcommand{\user}{Ryan\xspace}

\usepackage{mathptmx}                  % use matching math font

\begin{document}

%%%%%%%%%%%%%%%%%%%%%%%%%%%%%%%%%%%%%%%%%%%%%%%%%%%%%%%%%%%%%%%%
%%%%%%%%%%%%%%%%%%%%%% START OF THE PAPER %%%%%%%%%%%%%%%%%%%%%%
%%%%%%%%%%%%%%%%%%%%%%%%%%%%%%%%%%%%%%%%%%%%%%%%%%%%%%%%%%%%%%%%


\maketitle

\input{Sections/Intro}
\input{Sections/Related_Work}
\input{Sections/Design_Challenges_and_Goals}

\input{Sections/Method}
\input{Sections/Interface}
\input{Sections/Usage_Scenario}
\input{Sections/Discussion_and_Future_Work}

\bibliographystyle{abbrv-doi-hyperref}

\bibliography{template}

\end{document}

%% file: Sections/Intro.tex
\section{Introduction}

Fueled by advancements in deep learning (DL), especially transformer-based models, generative AI (GenAI) has rapidly gained prominence both in terms of technical sophistication and popularity. In the image generation domain, popular models such as DALL·E 3~\cite{dalle3} and Midjourney~\cite{midjourney} are now capable of creating high fidelity, detailed, and realistic images that easily deceive human viewers~\cite{cao2023comprehensive}. This emergence coincides with a rising demand for effective detection and analysis of realistic-but-fake images, as their misuse could lead to significant issues such as the misinformation spread~\cite{combating2023Xu}, copyright violations~\cite{sag2023copyright}, digital forensics challenges~\cite{guleria2023chatgpt}, and credit–blame asymmetry~\cite{porsdam2023generative}. This is clearly highlighted by recent regulatory initiatives, such as the E.U.'s AI Act addressing risks in AI~\cite{novelli2023taking} and a U.S. Executive Order on Safe and Trustworthy AI~\cite{biden2023executive}, which emphasize a critical need to develop new tools and techniques to support trustworthiness and authentication of potential GenAI-digital content.

More broadly, professionals that work with digital content, including  misinformation researchers, digital forensic analysts, social media moderators, and legal and copyright compliance experts, must increasingly identify and differentiate AI-generated content~\cite{elkhatat2023evaluating, hamed2024safeguarding}. One strategy, primarily focused on by the AI research community, has been to develop automated methods for detection; these have been shown to surpass human experts in distinguishing between AI-generated and human-created media (particularly when the data aligns with their training datasets)~\cite{ha2024organic}. Unfortunately, relying solely on automated detection tools imposes three significant challenges, as current approaches tend to lack (i)~generalizability~\cite{zhang2019detecting}, (ii)~interpretability~\cite{lloyd2023there, weber2023testing, sadasivan2023can}, and (iii)~accessibility. 

The challenge to generalization arises from the distinct visual signatures that different generative models imprint on their outputs, which vary between models (e.g., GAN models may create different visual signatures compared to diffusion models)~\cite{corvi2023detection}. Put another way, a detection method effective against one type of model may fail against another, and the constant introduction of new models exacerbates this problem~\cite{zhang2019detecting}.
% Despite the limitations of automated detection, its potential augment the expertise of human professionals in identifying and understanding AI-generated content is considerable. Recent research~\cite{ha2024organic} indicates that while automated tools may face challenges with generalization, they can outperform human experts (artists) in identifying AI-generated versus human-created art, provided the data is similar to their training data. Integrating automated detection with human analysis emerges as a particularly effective method for discerning and comprehending AI-generated images patterns, a strategy endorsed by the authors of these studies~\cite{ha2024organic, lin2024detecting}. Similarly, in AI-generated text detection, there is a call for the development of interpretable, interactive tools~\cite{weber2023testing}. Such tools would assist human professionals by summarizing patterns specific to AI-generated text, thereby enhancing their ability to detect it. 

One way to potentially surpass the limits of fully automated approaches is to integrate them into human-in-the-loop (HITL) workflows as a way to enhance the proficiency of humans in discerning and understanding AI-generated images and GenAi models~\cite{ha2024organic, lin2024detecting}. Recent advocacy in this area has likewise highlighted the need to create interpretable tools and techniques for these tasks~\cite{weber2023testing}, which could provide particular value to for users and stakeholders without technical background expertise in DL~\cite{lin2024detecting}. Building on the broader community efforts that employ HITL approaches for explainable AI and human-AI interaction~\cite{xu2023transitioning}, a solution here could aid human users by highlighting subtle (or latent) patterns unique to AI-generated content in human-interpretable ways, supporting the interrogation of fake image content and GenAI models. 

The primary motivation of this project is thus to help address the three challenges discussed above. To accomplish this, we develop and evaluate a first-of-its-kind HITL pipeline for the analysis of GenAI models and content. Our approach, implemented as a visual analytics tool called \sys, helps individuals who engage with AI-generated content to efficiently identify, analyze, and summarize patterns that suggest authenticity in AI-generated images. To the best of our knowledge, \sys is the first pipeline and interactive system that facilitates such analysis. Our contributions are summarized in high level as follows:

\begin{itemize}[leftmargin=*]

\item \textbf{\sys, a unified workflow and system}: \sys implements an end-to-end pipeline for efficient detection of AI-generated images and subsequent exploration, analysis, and summarization of their patterns. Our framework flexibly extends to images produced by models such as ProGAN~\cite{gao2019progan}, styleGAN~\cite{karras2019style} and the latent diffusion models~\cite{rombach2022high}.

\item \textbf{An image encoder to distill image authenticity/artificiality into compressed embeddings}: \sys develops an effective detector on top of large-scale pretrained models to classify real from AI-generated images. This detector's feature space serves as an image encoder, transforming images into compressed representations that are rich with latent indicators of their authenticity or artificiality. These representations are then utilized to discover patterns indicative of an image's realness or fakenss.

\item \textbf{An interpretability technique for pixel importance in image authenticity}: \sys applies gradient-based methods to identify crucial pixel groups that significantly impact the trained detector's real / fake prediction. This process uncovers recurring patterns in AI-generated images, particularly focusing on those deceptive patterns prevalent in misclassified fake images that account for incorrect predictions by the detector. Furthermore, it facilitates the identification of patterns in real images that cause them to be inaccurately flagged as fake, highlighting potential weaknesses in the detector.

\item \textbf{An efficient workflow for analyzing patterns}: \sys employs coordinated visualizations to support comparative analysis of patterns in AI-generated images and real images at various detail levels. These include a representation overview with novel cell glyphs, designed to enhance the exploration of real/fake patterns in a large number of images, an image view for displaying images with similar fake patterns as well as summarizing common visual patterns in these images, a dimension view for connecting quantitative patterns in the distilled representations with quantitative visual pixel patterns, and finally a pattern view that for detailed investigation into fake patterns in images with quantitative assessment to pixel groups' influence.

\item \textbf{Validation via usage scenarios}: We demonstrate the usefulness of our tool through two scenarios, showcasing how \sys can help discover fake patterns from GAN models such as proGAN and how it can be used to understand traits of human face images generated by diffusion-based deepfake technology. 
\end{itemize}

We conclude the paper with a discussion of lessons learned, including how HITL, visualization-driven approaches like \sys can provide generalizable insights to the community and open opportunities for new future work in this area.

%% file: Sections/Related_Work.tex
\section{Related Work}

To develop \sys, we build upon prior research regarding the detection of AI-generated images, the explanation of transformer models, and the design principles for visual analytics in explainable AI tasks:

\begin{figure*}[htbp!]
    \centering
    \includegraphics[width=0.9\textwidth]{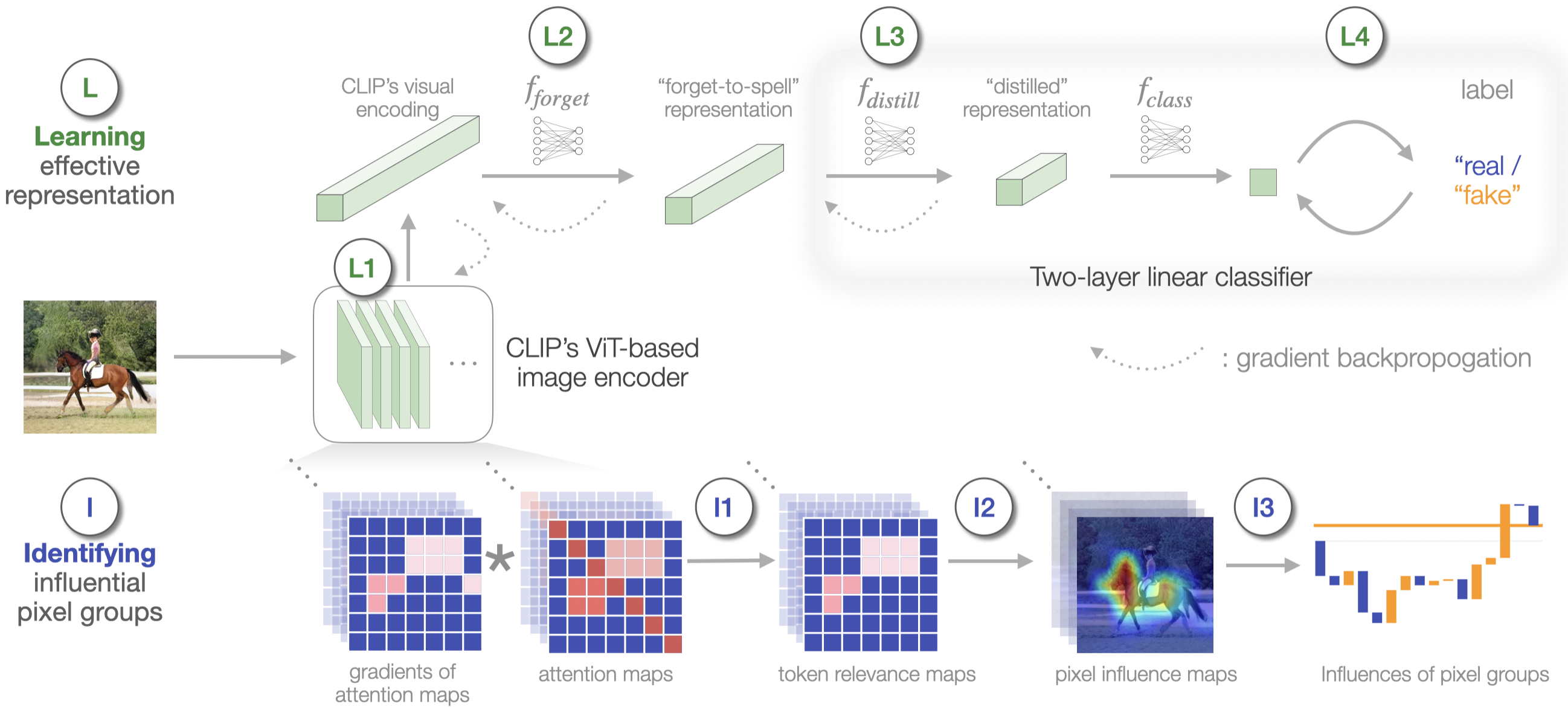}
    \caption{We employ a two-step approach for detecting AI-generated images and identifying their underlying patterns. Step L (Learning) begins with encoding images using CLIP's image encoder (L1) and then proceeds to strip textual information via a 'forget-to-spell' projection (L2). Subsequently, we train a classifier that includes a distiller layer, which maps the 'forget-to-spell' representation to a distilled space (L3), and a classification head that predicts the authenticity of images (real or fake) based on this distilled representation (L4). In Step I (Identification), we focus on identifying influential pixel groups by computing token relevance using a gradient-based technique (I1), translating token maps to individual pixel maps (I2), and computing uniform contribution metrics for pixel groups based on the classifier's weights (I3).}
    \label{fig:method}
\end{figure*}

\subsection{Detecting Fake Images}

Even long before the debut of generative image models (e.g., GANs were first proposed in 2014~\cite{goodfellow2014generative}), there existed a need to detect fake or modified images. Traditional image manipulation methods, such as via image editing software like Photoshop, were known to introduce detectable changes in image statistics, identifiable through visual cues including compression artifacts~\cite{agarwal2017photo}, resampling~\cite{popescu2005exposing}, or unnatural reflections~\cite{o2012exposing}. 
Early machine learning efforts at fake/manipulated image detection employed learning-based approaches~\cite{cozzolino2015splicebuster, rao2016deep, wang2019detecting}. With the rise of GenAI technologies, there has been a concerted parallel effort to devise classifiers capable of detecting images crafted by specific generative models~\cite{frank2020leveraging, marra2018detection, rossler2019faceforensics++}. 
However, these classifiers often encounter challenges in generalizability, particularly when confronted with out-of-distribution (OOD) data or newer models containing dissimilar architectures or methodologies~\cite{cozzolino2018forensictransfer, zhang2019detecting}. This challenge has motivated the development of more versatile classifiers designed to accommodate a broader spectrum of generative models~\cite{nataraj2019detecting, wang2020cnn, chai2020makes, asnani2022proactive, Ojha_2023_CVPR, zhu2023gendet}.

The main obstacle for learning-based method is to discover a feature space that is sufficiently robust to distinguish accurately between authentic images and a diverse array of AI-generated images, thereby ensuring reliable detection across various models ~\cite{zhu2023gendet}. Recent research~\cite{Ojha_2023_CVPR} has suggested employing the feature space from large pre-trained models (like CLIP) to take advantage of their comprehensive image encoding capabilities to detect AI-generated images. This method has shown to be effective and generalizable in subsequent studies~\cite{cozzolino2023raising}. 
We directly build upon these findings by modifying CLIP's visual encoder to develop a linear classifier that serves as an efficient image encoder. This encoder transforms images into a compact representation, where each dimension encodes distinct yet crucial information to the image's authenticity. These representations facilitate the identification of distinctive patterns within AI-generated images.

\subsection{Adapting CLIP for Downstream Tasks}
CLIP~\cite{clip} (Contrastive Language-Image Pre-training) has emerged as a powerful pre-training paradigm, seamlessly combining textual supervision with visual cues and demonstrating remarkable adaptability and success across a range of semantic tasks that include image segmentation~\cite{li2022language, rao2022denseclip, luo2023segclip}, object detection~\cite{bangalath2022bridging}, classification~\cite{zhou2022conditional} and medical imaging~\cite{zhao2023clip}. 
Unfortunately, the entanglement of text and vision poses challenges for tasks that are predominantly visual, as recent studies have highlighted that CLIP's visual encoder may inadvertently embed textual information into its outputs~\cite{materzynska2022disentangling}. Such unintended textual embeddings can compromise the efficacy of gradient-based methods aimed at identifying crucial non-semantic pixels, consequently affecting the accurate identification of inauthentic image regions.

For \sys, we overcome this by creatively integrating a ``forget-to-spell'' model: an orthogonal linear projection designed to minimize textual content in CLIP's latent space. This approach successfully removes unwarranted textual embeddings, yielding a predominantly visual-centric representation that better serves our task of fake image detection and fake pattern discovery.

\subsection{Measuring Token Influence In Transformer Models} 
As the transformer architecture becomes increasingly central to modern deep learning, there is a growing emphasis on explaining transformer models ~\cite{barkan2021grad, ali2022xai, chefer2021generic, bracsoveanu2020visualizing, yuan2021explaining}.  A notable method for interpreting transformers is by attributing influence scores to individual tokens~\cite{chefer2021generic, barkan2021grad}, aiming to measure each token's importance in the model's predictive outcomes. The vocabulary surrounding this idea varies with terms like ``relevance,'' ``importance,'' and ``influence'' being used interchangeably~\cite{yuan2021explaining}.

In the context of vision transformers, where tokens correspond to image patches, computing token relevance is essential for identifying the pixels that most significantly impact the model's decision-making ~\cite{barkan2021grad}. With a trained real/fake image detector, revealing influential pixels to predictions is key to identifying the patterns of AI-generated images.

To accomplish this, we adapt the relevance map computation technique introduced by Hila et al.~\cite{chefer2021generic} based on its compatibility with encoder-decoder architectures like CLIP, which serves as the basis for our image encoder. Specifically, we modify its relevance update rules to align with our task, and utilize the gradients propagated back from the classifier's distiller layer to the final transformer block of CLIP's visual encoder. These gradients are used as coefficients to aggregate attention maps from the block's multiple attention heads, producing a relevance map. This map effectively highlights the pixels that are crucial for determining image authenticity, thereby exposing patterns indicative of AI-generated images.

\subsection{Visual Analytics for Generative AI}
Visual analytics systems play a crucial role in augmenting various human-AI tasks, including data augmentation~\cite{zhao2021human}, labeling~\cite{hoque2022visual, jia2021towards}, data exploration and analysis~\cite{wang2023wizmap,wang2023drava, zhang2022sliceteller, gou2020vatld, park2021vatun, kwon2018retainvis}, bias inspection and mitigation~\cite{ahn2019fairsight,ghai2022d,xie2021fairrankvis,kwon2022dash,kwon2023finspector}, model explanation~\cite{wang2020cnn, wang2018dqnviz, mishra2022not, huang2022conceptexplainer}, and model refinement~\cite{wongsuphasawat2017visualizing, huang2023intervls}.

With the rapid progression of GenAI, there is a burgeoning interest in research focusing on visual analytics for generative models. Recent work includes visualizing transformer attention mechanisms~\cite{yeh2023attentionviz, gao2023transforlearn, park2019sanvis}, illustrating diffusion processes~\cite{lee2023diffusion}, and enhancing prompt engineering~\cite{mishra2023promptaid,feng2023promptmagician}. The primary focii of visual analytics tools in the context of generative AI (vis4GenAI) is to employ visualization techniques to unravel the inner workings in generative models,  particularly those grounded in transformer or diffusion methodologies and to support interactions with generative models like prompt engineering.

Despite these advances, there exist open gaps in in understanding the characteristics of generative model outputs (a field rapidly gaining prominence). To help address this, \sys is, to the best of our knowledge, the first system employing visual analytics to deeply scrutinize and comprehend patterns in AI-generated images.

%% file: Sections/Design_Challenges_and_Goals.tex
\section{Design Challenges and Goals}
Our goal is to create an interactive system that helps users analyze the patterns of AI-generated images. This goal faces two primary challenges: (i) We need to devise a method to broadly, qualitatively, and quantitatively extract patterns that distinguish fake images from real ones; (ii) We need to develop an accessible (i.e., user-friendly) interface that facilitates in-depth analysis of these patterns at varying degrees of detail. These primary obstacles are subdivided into specific challenges, labeled as \textbf{C1}-- \textbf{C4}.

\subsection{Design Challenges}
\begin{itemize}[leftmargin=*]

\item \textbf{(C1) Automatic Discovery of Fake Patterns}: In scenarios where classifiers are trained to differentiate between real and AI-generated images, they inherently identify unique patterns of both types \cite{cozzolino2023raising}. The challenge is to devise a way to utilize these classifiers to efficiently extract these distinct patterns amidst irrelevant visual information that might confound the analysis. 

\item \textbf{(C2) Quantifying  Influence of Fake Patterns}: The diversity inherent in generative models leads to a wide range of artificial patterns, even within outputs from the same model \cite{corvi2023detection}. A key to analyzing these patterns is to distinctively identify and quantitatively assess their impact. However, existing techniques for evaluating pixel significance tend to focus on individual pixels rather than the collective influence of pixel groups that form a pattern ~\cite{park2022vision}. Furthermore, there are no uniform metrics for significance measurement that are comparable across multiple images.

\item \textbf{(C3) Efficient Analysis and Summarization of Fake Patterns}: After identifying fake patterns constituted by influential pixel groups, the subsequent challenge is to devise an interface enabling users to effectively examine and summarize these patterns. The primary obstacle is the potential for an overwhelming number of patterns \cite{wang2023wizmap}, which could hinder clarity in analysis. At current, there are no established guidelines about how to design such an interface or tailor the user experience for such a workflow.

\item \textbf{(C4) Adapting to a Diverse Range of Generative Models}: As GenAI technology rapidly evolves, it is critical that the analysis system adapts across various models. The system's design must also be flexible enough to accommodate new models as they emerge \cite{Ojha_2023_CVPR,cozzolino2023raising,zhu2023gendet}, ensuring its usefulness over time. This requires a modular approach, the specifics of which are yet to be determined.

\end{itemize}

\subsection{Design Goals}
\label{subsect:design_goals}

To address the design challenges \textbf{C1}--\textbf{C4}, we outline a set of six primary design goals \textbf{G1}--\textbf{G6}. The design and implementation of \sys is intended to support these goals.

\begin{itemize}[leftmargin=*]

\item \textbf{(G1) Develop a classifier to detect real from fake images using large-scale pre-trained models}. The process of discovering fake patterns \textbf{(C1)} in images involves two steps: (i) Learning to distill critical information that separates real from fake images into a condensed representation. (ii) Leveraging this representation to detect key pixel groups and thus identify artificial patterns. For the initial step, we employ a supervised learning strategy, which is currently considered the state-of-the-art~\cite{lin2024detecting}. Training a classifier to discern between real and fake images indirectly enables it to encode pixel information that can indicate an image's authenticity \cite{Ojha_2023_CVPR, cozzolino2023raising}. This trained classifier also needs to be able to extract features of artificiality from a wide range of AI-generated images \textbf{(C4)} produced by various models. Our strategy involves building this classifier on a pre-trained model known for its extensive training on a varied dataset, ensuring exposure diverse image distributions. We freeze the underlying pre-trained model to capitalize on its generic pixel information encoding capabilities, subsequently developing classifiers on top of its visual encodings.

\item \textbf{(G2) Using gradients to identify important pixels}. In addressing the second step (identifying critical pixel groups), we leverage gradients. Gradients, which are propagated back from the ``distilled'' representations, reveal the sensitivity of the output (distilled representation) to changes in the input (images) \cite{chefer2021generic}. By basing our classifier on a pre-trained CLIP model, we focus on the gradients directed towards the attention map within its transformer block. This process uncovers token sensitivities, which are then mapped back to pixel sensitivities, effectively identifying pivotal pixels \textbf{(C1)}. Given the multi-dimensional nature of our distilled representation, we can craft detailed masks for each dimension, as each dimension uniquely encodes information about the image's realness or artificiality. This technique allows for the creation of comprehensive masks that trace specific pixel groups affecting the image's authenticity.

\item \textbf{(G3) Establish a uniform metric for evaluating pattern influence}. Following the identification of patterns, quantifying their influence becomes essential \textbf{(C2)}. The gradient-based relevance maps used to identify significant pixels fall short for comparative purposes due to their focus on relative importance within single images ~\cite{chefer2021generic}. To overcome this, we utilize the trained classifier's weights. Each identified pattern correlates to a representation dimension, possessing an associated weight within the classification head and a corresponding representation value. The weight reflects the pattern's global significance, while the value offers local context. Multiplying these factors yields a scalar that integrates both perspectives. With a subsequent normalization, we achieve a uniform metric for assessing the influence of patterns across and within images.

\item \textbf{(G4) Facilitate counterfactual analysis}. Counterfactual reasoning plays a pivotal role in providing in-depth explanations and insights into data \cite{akula2022cx}, particularly in vision tasks such as analyzing fake image patterns \textbf{(C3)}. To support this, we aim to enable users to conduct comparisons between images that are correctly classified due to their apparent traits and those that are misclassified because of their subtle, deceptive characteristics. In addition, we support ``what if'' scenarios \cite{mishra2022not, stepin2021survey} as a way to investigate the changes required for a misclassified AI-generated image to be correctly identified, as a way to support in-depth understanding of the key factors contributing to misclassification.

\item \textbf{(G5) Enhance understanding through comparative analysis} Another effective way to enhance understanding of fake/real patterns  \textbf{(C3)} is comparative analysis; that is, by organizing image data points based on their distilled representations, users can identify patterns and similarities more effectively ~\cite{stepin2021survey}. Presenting misclassified images alongside their distilled representations further allows users to draw connections between qualitative patterns and quantitative traits, enhancing their insights. Moreover, we propose that allowing users to label unique pattern cells facilitates easier navigation and reference during analysis.

\item \textbf{(G6) Enable detailed examination of individual images.} For a thorough analysis and summarization of artificial patterns \textbf{(C3)}, it is essential to allow users to examine individual images. This detailed inspection enables them to trace and review the origins of fake patterns directly. By supporting granular comparisons, such as investigating a few selected images to uncover the influential pixels related to authenticity, users can achieve a more detailed and nuanced understanding of the patterns present in a subset.

\end{itemize}

% \begin{figure*}[htbp!]
%     \centering
%     \includegraphics[width=\textwidth]{figs/violinplot.png}
%     \caption{The user conducts a comparison between the dimension contributions (H2 and H5) of two "brown/black horse head" cells, specifically one with severe sensitivity (H1) and another with minor sensitivity (H4). The analysis uncovered a notable difference in dimension 2 across these cells. Further investigation into the pixel groups associated with dimension 2 revealed a strong correlation with pixels adjacent to or resembling horse eyes (noted in H3 and H6). It was observed that images in the severe sensitivity cell exhibit high values in dimension 2, whereas those in the minor sensitivity cell show low values. This pattern indicates that the realistic portrayal of horse eyes might significantly influence the detection between real and AI-generated horse head images.}
%     \label{fig:violinplot}
% \end{figure*}

%% file: Sections/Method.tex
\section{Representation Learning and Pixel Contribution Measurement}
\subsection{Real and AI-Generated Datasets}

In this paper, we evaluate the efficacy of \sys through the application of two large state-of-the-art datasets, each composed of AI-generated images originating from two leading generative model architectures: Generative Adversarial Networks (GANs) and diffusion models.

To assess the system's proficiency in analyzing images synthesized by GANs, we employ the proGAN dataset~\cite{wang2020cnn}. This dataset includes authentic images from the PASCAL object detection challenge, which includes 20 classes with their fake counterparts produced by the proGAN model. This dataset presents a balanced composition, with 30,000 real and 30,000 fake images per class.

For a diffusion-based model, we use the DetectingSyntheticImage dataset~\cite{corvi2023detection}, which focuses on images fabricated by latent diffusion models (LDMs). This dataset is particularly diverse, featuring diffusion-generated images across different categories such as ImageNet classes, architectural subjects (including bedrooms and churches), and synthetic human faces. Specifically, we examine the 20,000 fake human face images generated by LDM in the dataset, together with their corresponding real human face images from the Flickr-Faces-HQ (FFHQ) dataset, which consists of 70,000 high-quality human face images.

For each pair of real and fake images -- such as real horse images versus those synthesized by proGAN -- we train a classifier to discern between ``real'' and ``fake,'' using the trained classifier to identify deceptive patterns among the generated images.

% \subsection{Method}
\subsection{Learning Effective Representations}
For an image $I$, the system begins by preprocessing it using the CLIP:ViT-B/32' image encoder, denoted as $f_{\text{CLIP}}^{image}(\cdot)$. This step encodes the image into a 512-dimensional vector representing generic visual features (L1 in Fig.~\ref{fig:method}). To refine this representation, a ``forget-to-spell'' projection, $f_{\text{forget}}(\cdot)$, is employed to filter out any unintended text-related information that might have been incorporated into the image representation $f_{\text{CLIP}}^{image}(I)$, during the encoding (L2 in Fig.~\ref{fig:method}). This results in a more visually focused 256-dimensional vector $(f_{\text{forget}} \circ f_{\text{CLIP}}^{image})(I)$, which retains essential visual features of the image.

Following this, the system employs a linear classifier trained on the forget-to-spell representation to discern between genuine and synthetic images. This classifier is composed of a distiller layer, $f_{distill}(\cdot)$, which concentrates the representation by learning critical orthogonal dimensions, thus reducing the 256-dimensional forget-to-spell vector to a more manageable 16-dimensional vector (L3 in Fig.\ref{fig:method}). The subsequent classification head, $f_{class}(\cdot)$, leverages this distilled vector to classify the image as real or fake (L4 in Fig.\ref{fig:method}).

Upon the classifier's training completion, the combination of $f_{CLIP}^{image}()$, $f_{forget}$ and $f_{distill}$ forms an effective image encoder. This encoder converts images into a compact space rich with information crucial for determining image authenticity, thereby facilitating subsequent pattern discovery \textbf{(G1)}. This details are explained as follows:

\subsubsection{Removing Latent Text Information}
Research indicates that CLIP's visual embeddings, $f_{\text{CLIP}}^{image}(I)$, provide vital hints for differentiating between authentic and synthesized images~\cite{Ojha_2023_CVPR, cozzolino2023raising, zhu2023gendet}. However, due to the nature of CLIP's contrastive learning process, there is an inherent mixing of visual and textual information within its embeddings~\cite{materzynska2022disentangling}. While the blending of text and visual features benefits tasks requiring an understanding of pixel semantics (e.g., zero-shot classification, vision question answering), it presents obstacles for tasks focused on verifying image authenticity, which rely primarily on visual data. The inclusion of additional text information may obscure the effectiveness of gradient-based methods in identifying key pixels, resulting in less accurate detection of spurious influential regions in images.

To counteract this issue, the forget-to-spell model, $f_{forget}(\cdot)$, equipped with an orthogonal projection technique~\cite{materzynska2022disentangling}, is employed to remove textual data from CLIP's embeddings. This model is trained with an orthogonality loss function to prevent the overlap of visual and text representations in the projection space. It is trained on a diverse dataset including natural images with corresponding labels, text-only images, and images featuring overlaid text (with both meaningful and nonsensical strings). This training regimen is designed to instruct the model in distinguishing visual elements from textual content within CLIP's embeddings, thus reinforcing its focus on visual details. The outcome is a visual-centric representation, $(f_{forget}\circ f_{CLIP}^{image})(I)$, ready to be used for vision-only tasks.

\subsection{Pixel Contribution Measurement} 
To compute the impact of image pixels in determining its authenticity, gradients from each of the final distilled representation's 16 dimensions back to CLIP's transformer block attention maps, $\frac{\partial A}{\partial x_{i}}, i\in{1,...,16}$, are calculated. This generates 16 detailed maps highlighting pixel groups that significantly sway the classification of the image as real or fake \textbf{(G2)}. By employing the global weight of the classifier, $f_{class}$, a uniform approach to evaluate the pixel groups’ influence is established \textbf{(G3)}. These maps also facilitate the segmentation and clustering of these pixel groups into visual concepts, enhancing the efficiency in analyzing common fake patterns among images\textbf{(G5)}. The discovered patterns are then wrapped up in an interactive interface for users to analyze and summarize (Section~\ref{sect:interface}). The detailed implementation of this process is as follows:

\subsubsection{Distilling Critical Information}
\label{subsection:distill}
After removing textual information with the forget-to-spell model and obtaining 256-dimensional vectors $f_{forget}(f_{CLIP}^{image}(I))$, we obtain visual-centric representations. But they still generically contain all kinds of visual information, so we need to further distill from them critical information that indicates image reality/fakeness.

To this end, we employ a supervised learning approach, essentially training a classifier on the forget-to-spell representations to perform binary classification: determining whether an image is real or fake. This classification process condenses crucial information for authenticity detection into a more manageable feature space. By analyzing the gradients of these representations, we can identify specific pixels in the original images that significantly impact the model's predictions.

Our classifier is deliberately simple, comprising a two-layer linear model. The first layer, the distiller, $f_{distill}(\cdot)$, reduces the dimensionality of the forget-to-spell representations into a smaller subspace. The second layer, acting as a classification head, $f_{class}(\cdot)$, uses these distilled representations to predict a binary outcome, which is then processed through a sigmoid function to produce the final classification label.

Training the classifier involves using a dataset containing both real and fake images, alongside their corresponding labels (the Universal Fake Detect dataset in our case~\cite{Ojha_2023_CVPR}). Once trained, the classifier can efficiently distill the forget-to-spell representations into a critical space, significantly reducing dimensionality while preserving information essential for distinguishing between real and fake images. This process yields an encoder that effectively highlights crucial details for authenticity verification.

During training, we ensure that the distiller's weights are initialized orthogonally and incorporate an orthogonality penalty in the loss function. This encourages the learning of a space defined by uncorrelated dimensions, maximizing the retention of information. Our loss function is defined as $Loss(y,\hat{y}) = \lambda_{BCE} BCE(y,\hat{y}) + \lambda_{ortho} R(W)$, , where $y$ is the image label, $\hat{y}$ is the classifier's output, and $R(W) = \|I - WW^T\|_F^2$ is the orthogonality penalty for the distiller's weight $W$.  The hyperparameters $\lambda_{BCE}$ and $\lambda_{ortho}$ are empirically set to 3 and 1 respectively to optimize classifier performance, and the model is trained using the Adam optimizer, with a batch size of 32 and a learning rate of 1e-3. The training is done in a MacBook Pro Laptop with M1 Chip.

With the model properly trained, we can condense the visual-centric yet generic forget-to-spell representations into a compact representation enriched with critical information for image reality/fakeness. Using these condensed vectors, we then apply a gradient-based method to highlight influential pixels and quantify their influences.

\subsubsection{Identifying Influential Pixel Groups}
\label{subsection:relevance_map}
We modify a method from ~\cite{chefer2021generic} to compute a pixel relevance map for identifying relevant pixels to the final real/fake prediction. We then use weights learned in the classification head $f_{class}(\cdot)$ to quantify each pixel group's contribution. The map is computed based on gradients of attention maps in the transformer blocks within CLIP. 

Recall how ViT works: for an image $I$, ViT divides it into $k \times k$ patches (with $k = 224 / 32 = 7$ for the CLIP:ViT-B/32 architecture). These patches are then tokenized into vectors, resulting in a total of $k^2$ tokens per image. An additional [CLS] token is concatenated, forming a $(k^2 + 1) \times d$ matrix, where $d$ is the embedding space dimension (512 in this case).

To derive the pixel relevance for an image of shape $h\times w $, we first compute a token relevance map—a $k \times k$ matrix indicating each token's influence on the prediction. This token map is then upscaled to the original image size through interpolation and normalized between 0 and 1 to facilitate interpretation.

The computation of token relevance, $R_{k\times k}$, for the CLIP:ViT-B/32 model involves gradients from the attention maps within the transformer blocks. Initially, all tokens are considered equally contributory, represented by a matrix where every element is 1, $[1]_{k\times k}$. 

\begin{equation}
R =[1]_{k\times k}
\end{equation}

We then refine this relevance map by backpropagating from the last transformer block, updating the relevance map $R$ by aggregating updates using the rule:

\begin{equation}
R = R + \bar{A_{i}} R, i \in {m,...,1}
\end{equation}

Here, $R$ is the relevance map and $\bar{A_i}$ is derived from gradients of attention maps in the $i$-th transformer block (suppose we have m blocks in total), emphasizing elements of with a significant impact on the output—both in terms of the magnitude of the attention and its sensitivity to changes. $\bar{A_i}$ is computed as:
\begin{equation}
\bar{A_{i}} = \mathbb{E}_h\left( \nabla A_i \odot A_i \right)^+
\end{equation}

where $\nabla A_i$ represents the gradient of the attention map $A_i$, and the operation $\odot$ denotes the Hadamard product.

However, aggregating attention maps from all transformer blocks may dilute the information with less pertinent details. Thus, to maintain focus on the most influential pixels, we only consider the relevance information from the last transformer block. This ensures that the computed relevance map specifically highlights the pixels with the highest relevance to the authenticity of the image. The final relevance computation is simplified to:

\begin{equation}
\label{eq:relevance}
R = \mathbb{E}_h\left( \nabla A_{last} \odot A_{last} \right)^+
\end{equation}

This method involves running a forward pass to obtain both the distilled representation and the attention maps, followed by calculating the gradient of the attention maps. The relevance map is then computed according to the formula Eq.~\ref{eq:relevance} above and resized to match the original image size, serving as a mask to identify and highlight image regions crucial for distinguishing real images from fake ones.

\subsubsection{A Uniform Metric for Pattern Contribution}
\label{subsect:pixel_contribution}

For each image, we produce 16 distinct relevance maps. Each map is derived from one of the 16 gradients corresponding to the dimensions in the distilled representation, which encodes clues about the image's authenticity. These maps precisely highlight specific pixel groups pivotal to the authenticity prediction. However, the relevance scores within each map are only meaningful within the context of that specific map. Direct comparison between maps or using the relevance scores to assign a universal influence score to the pixel groups is not sensible due to this relative nature.

To overcome this limitation, we introduce a methodology that integrates both global and local dimensional information to calculate a scalar value, normalized between -1 and 1. This scalar's sign indicates the direction of the pattern's contribution (positive for fake, negative for real), while its magnitude reflects the extent of the contribution.

The process for calculating this contribution value is as follows: For an image $I$, employing the gradient-based method in Section~\ref{subsection:relevance_map}, we identify 16 pixel groups $pg_i, i \in {1,...,16}$, each linked to a dimension in the distilled representation, represented as $v_{distill} = (v_1, ..., v_{16})$. Using the classification head's weight, a 16-dimensional vector $w = (w_1, ..., w_{16})$, we calculate a scalar $s_i = v_i \times w_i$ for each dimension. This scalar, blending the local dimension value and the global dimension weight, becomes uniformly comparable across dimensions for the same image. For cross-image comparisons, normalization of $s_i$ is necessary. Initially, one might consider normalizing $s_i$ by dividing the sum $\sum_{i=1}^{16} s_i$ but this approach could inadvertently amplify the result due to potential cancellations in the summation of positive and negative values, leading to a smaller denominator. To circumvent this, we normalize $s_i$ into a contribution score $c_i$ as follows:

\begin{equation}
    c_i = \frac{s_i}{\sum_{i=1}^{16} |s_i|} = \frac{v_i \times w_i}{\sum_{i=1}^{16} |v_i \times w_i|}
\end{equation}

This formula ensures the preservation of the scalar $s_i$'s sign, vital for our analysis since the contribution's direction (negative for real and positive for fake) is crucial. The normalized score encapsulates all dimensional data, both local and global, for a pixel group, maintaining the original scalar's sign within the -1 to 1 range.

This method of measuring the contribution value facilitates an efficient comparison of the influence exerted by different pixel groups on the model's prediction of an image as real or fake within images and across images, providing a consistent metric to assess their impact.

\subsubsection{Revealing Commonality of Similar Images}
\label{subsect:visual_concept}

To illustrate commonalities among fake images exhibiting similar artificial patterns, we leverage concept-based methods~\cite{kim2018interpretability}. Initially, we utilize the relevance maps generated in Section~\ref{subsection:relevance_map} as masks to identify influential image segments (pixel groups) within a collection of similar images. These images are considered similar based on the proximity of their ``distilled'' representations in the L2 distance metric. To refine our analysis, we exclude overlapping segments within images by discarding those with high Intersection Over Union (IOU) scores relative to segments already identified. Subsequently, we encode these segments using the combined encoder $(f_{distill}\circ f_{forget} \circ f_{CLIP}^{image})(\cdot)$ into distilled representations. We group these representations into three clusters for analysis using k-means clustering. 

This process enables us to identify and highlight clusters of visual concepts, effectively revealing the common visual patterns shared by groups of images with similar authentic or fake patterns.

%% file: Sections/Interface.tex
\begin{figure}[t]
\centering
\includegraphics[width=\linewidth]{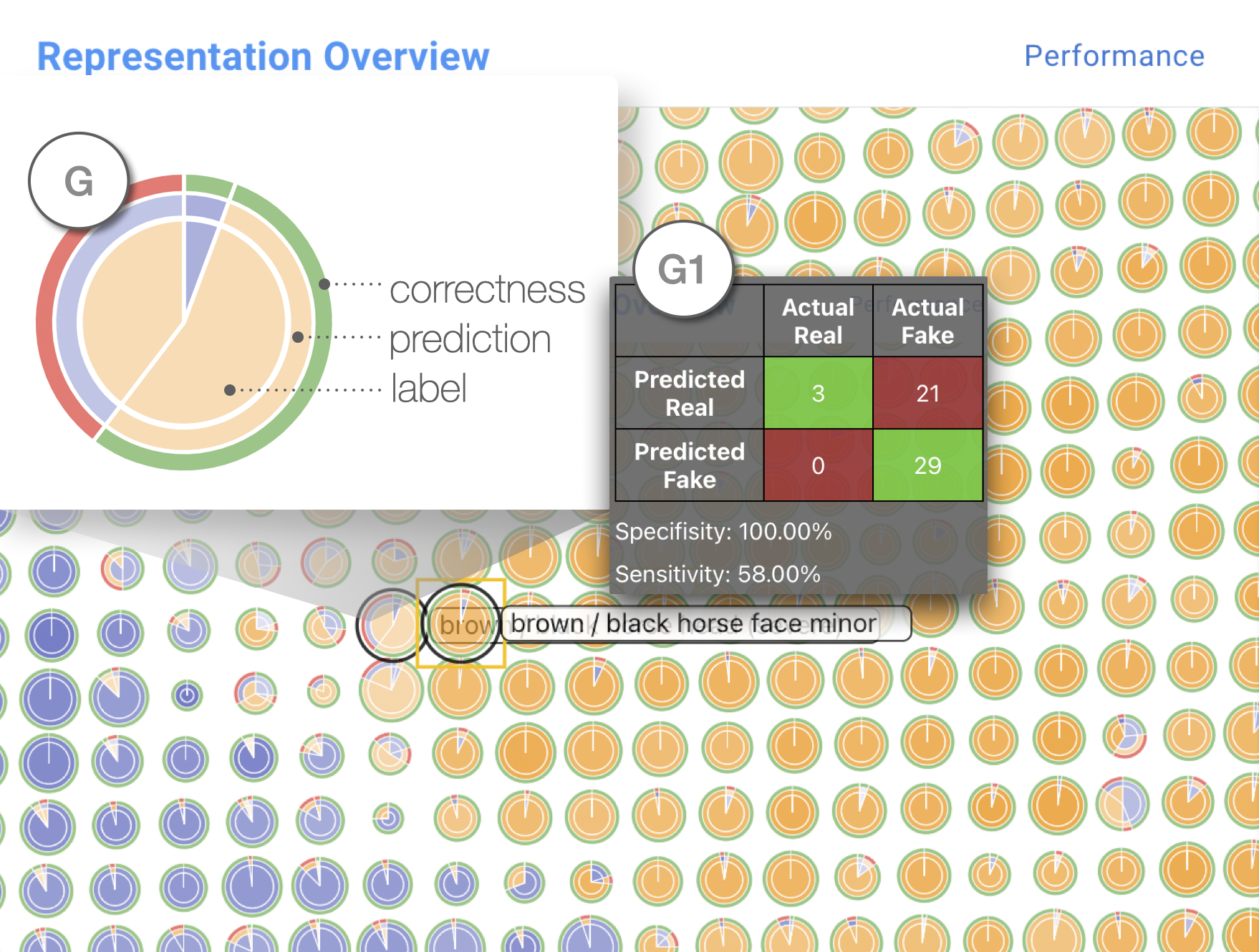}
\caption{Our system introduces a novel cell glyph (G) to aggregate and visualize information for images within a cell, facilitating navigation in the representation space. The user can easily identify fakeness severity levels by hovering on the cell (G1). In this instance, the user examines the ``brown/black horse head'' cell with severe sensitivity (see Fig.\ref{fig:violinplot}).}
\label{fig:cell_glyph}
\end{figure}

\section{The \sys Interface}
\label{sect:interface}

Building upon our design goals outlined in Section~\ref{subsect:design_goals}, techniques for extracting critical image authenticity information in Section~\ref{subsection:distill}, and methods for identifying influential pixel groups in Section~\ref{subsection:relevance_map}, we introduce \sys. This interactive system facilitates an interpretable analysis and summarization of fake patterns in images produced by various generative models, across multiple levels of data granularity (i.e., global, subset, and local levels).

Users begin their analysis by selecting a generative model and a specific subject for comparison (e.g., real versus fake ``horses'' generated by the ``proGAN'' model) using the header bar. The \sys interface consists of four main interactive components: the Representation Overview, the Image View, the Pattern View, and the Dimension View, each offering unique insights and interactions to explore fake image patterns.

\subsection{Representation Overview}
The Representation Overview (Fig.~\ref{fig:teaser}A) provides a summary of fake patterns, facilitates navigation and comparison, and supports user annotations by grouping similar image representations into distinct, non-overlapping cells while minimizing visual clutter \textbf{(G4)}. The objectives are mainly achieved through a custom cell glyph design which provides a summary of the fake image detector on the corresponding subset of images. The glyph intuitively presents the proportion of fake images within each cell that remain undetected by the fake image detector (Fig.~\ref{fig:cell_glyph}).

\textbf{Avoiding visual clutters via local aggregation.} Using the combined encoder $f_{distill} \circ f_{forget} \circ  f_{CLIP}^{image}$, we transform images into lower-dimensional distilled representations. This process latently places images with similar real or fake patterns close to each other in the distilled space. Traditional visualization techniques like t-SNE or UMAP, which compute 2-D coordinates for these representations, are not applicable directly due to their tendency to cause overlapping and congestion, leading to visual clutter. Instead, after using t-SNE to assign each representation a 2d coordinate, we aggregate the representations that are locally proximate (indicating similarity in realness or fakeness) by dividing the space into an $m \times m$ grid, with $m$ being a modifiable parameter set to 30 through empirical determination. This grid formation creates $m \times m$ non-overlapping cells and facilitates local aggregation.

\textbf{Enhancing deceptive pattern identification with cell glyphs.} A critical aspect of our visualization approach involves designing cell glyphs that highlight the presence of deceptive patterns in significant cells. Recalling our use of a classifier to distinguish real from fake patterns, it is noteworthy that instances misleading the classifier—such as fake images erroneously classified as real—contain particularly deceitful patterns. Thus, our cell glyph is designed to visually reveal the classifier's specificity and sensitivity, thereby indicating the severity of fakeness within cells.

Our glyph design, refined through multiple iterations, is a special pie chart divided radially into three layers (inner, middle, and outer) and further segmented into four sectors. These sectors represent true positives (fake images being correctly predicted), true negatives (real images being correctly predicted), false positives (real images being mispredicted), and false negatives (fake images being mispredicted). The outer arc of the glyph visualizes prediction results, with color coding maintained consistently across all views: real images in blue, fake in orange, correct predictions in green, and incorrect in red. The saturation level within each sector indicates the classifier's average confidence level, with paler colors denoting uncertainty or difficulty in pattern identification and more saturated colors indicating clearer, easily distinguishable patterns. To facilitate the understanding of the glyph the representation overview also enables a ``legend'' button that the user can click to show a legend depicting the meaning of each color-coded sector.

\textbf{Analyzing cells with similar images but diverging statistics} By deriving cells from their distilled representations, we guarantee that adjacent cells reflect akin real/fake information. This arrangement supports a robust comparative analysis framework. Exploring a cell and then examining its adjacent cells for similar images allows users to recognize differences in statistics between the cells, laying a strong foundation for comparative scrutiny.

\textbf{Select a cell to investigate images within.} The size of the glyph reflects the number of examples it contains, allowing users to quickly ascertain the prevalence of underlying patterns. This design enables users to identify challenging examples near the decision boundary at a glance, facilitating the transition to more detailed subset-level exploration. Selecting a cell highlights it with a gold square, and users can pan and zoom to navigate through the space.

\textbf{Annotation for incremental analysis.} For further engagement, users can annotate cells by right-clicking and type their description about observed patterns, with annotations visually marked and persisting until removed. This feature enhances the analytical nature of the analysis, empowering users to incrementally develop their understanding of fake image patterns.

\subsection{Image View}
To enhance the analysis of recurring fake patterns among similar images, we introduced the Image View (Fig.~\ref{fig:teaser}B). This view activates when a user selects a cell in the representation overview, displaying the images within to easily identify common patterns among fake images \textbf{(G5)}.

To ensure the images are organized in a visually intuitive manner, we utilize the IsoMatch method~\cite{fried2015isomatch}, which preserves the pairwise distances between images' t-SNE coordinates in a structured 2D grid layout. This approach prevents overlap within cells while maintaining proximity, facilitating a clearer analysis. 

Each image features a circle in its top-right corner that denotes its ground truth label —blue indicates real, and orange represents fake. Additionally, a drop shadow around an image indicates the prediction of the classification head, $f_{class}(\cdot)$. A discrepancy between the ground truth and the prediction, especially with fake images wrongly identified as real, indicates the necessity for in-depth analysis to identify deceptive patterns that fool the classifier.

\textbf{Toggle to see influential visual concepts.} The Image View also includes a ``Show Concept'' feature, which, when activated, displays a concept view highlighting influential visual concepts (See Fig.~\ref{fig:concept_smile}) among a group's images, using methods outlined in Section~\ref{subsect:visual_concept}. This feature aids in the users' understanding of a cell and fake patterns within. 

\textbf{Brushing to select images for further analysis.} For more in-depth examination, users can select multiple images using a brushing technique, gathering them into a ``Pattern View'' for a closer look at critical pixels and the fake patterns they contain.

\begin{figure*}[htbp!]
    \centering
    \includegraphics[width=.9\textwidth]{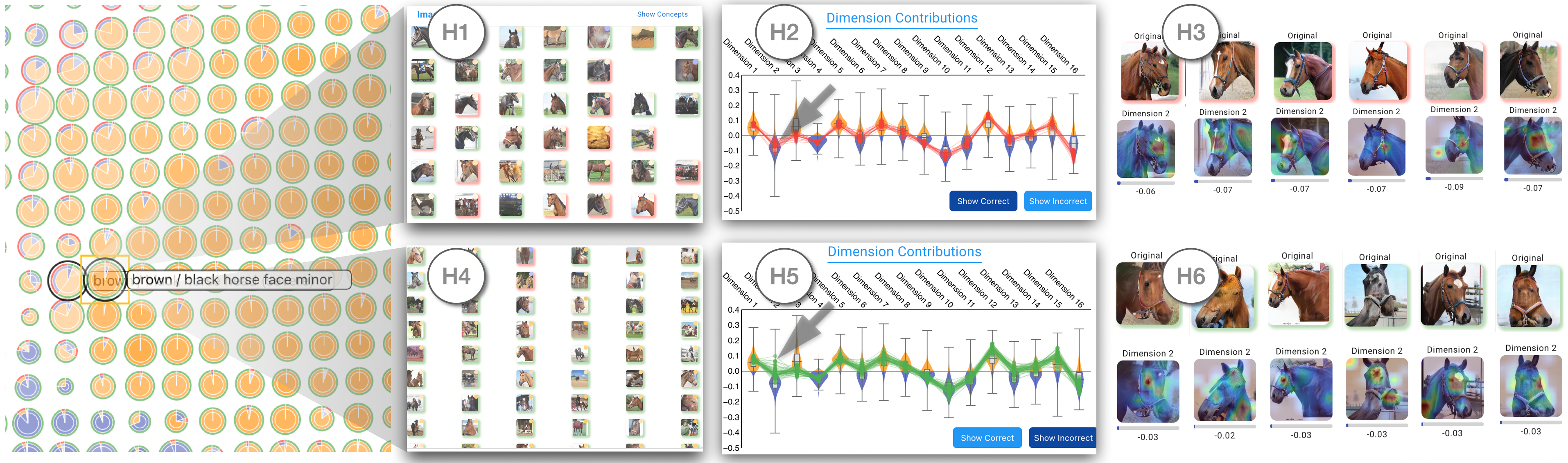}
    \caption{The user conducts a comparison between the dimension contributions (H2 and H5) of two ``brown/black horse head'' cells, specifically one with severe sensitivity (H1) and another with minor sensitivity (H4). The analysis uncovered a notable difference in dimension 2 across these cells. Further investigation into the pixel groups associated with dimension 2 revealed a strong correlation with pixels adjacent to or resembling horse eyes (noted in H3 and H6). It was observed that images in the severe sensitivity cell exhibit high values in dimension 2, whereas those in the minor sensitivity cell show low values. This pattern indicates that the realistic portrayal of horse eyes might significantly influence the detection between real and AI-generated horse head images.}
    \label{fig:violinplot}
\end{figure*}

\subsection{Pattern View}

To enable detailed examination of individual images \textbf{(G6)}, we introduce the ``Pattern View.'' (Fig.~\ref{fig:teaser} D) This view is activated when users brush to select images within the Image View for an in-depth analysis, subsequently updating to showcase the influential patterns of each selected image across distilled dimensions.

In the layout, we arrange all the information necessary for a detailed image examination in a single row. It begins with the original image, highlighted by a drop shadow to indicate the correctness of its prediction. Adjacent to the image, we display 16 unique heat maps. These maps represent the 16 diverse pixel relevance maps created to identify influential pixel groups within the image (see Section~\ref{subsection:relevance_map}). Additionally, we illustrate the contribution of these pixel groups (see Section~\ref{subsect:pixel_contribution}) — indicating their influence on the prediction —below each heatmap with a bar. An orange bar signifies a positive contribution, pushing the prediction towards 'fake,' while a blue bar means a negative contribution, pushing the prediction towards 'real'. For instance, in Fig.\ref{fig:teaser} D, dimension 10 shows a significant negative contribution to the images of a brown horse, indicating that the corresponding pixel groups misleadingly persuade the classifier to perceive those images as real.

At the end of the row, a waterfall chart showcases the contribution of each dimension to the image's perceived authenticity. This chart is particularly effective in revealing dimensions that exert substantial influence, thereby allowing users to easily pinpoint the most impactful patterns determining the image's authenticity or falseness. For example, the chart in Fig.~\ref{fig:teaser} D clearly identifies dimension 10 as a critical factor in misclassifying a fake image as real, facilitating straightforward identification by users.

At the row's end, a waterfall chart illustrates the contribution of each dimension to the image's perceived authenticity. The chart exposes dimensions that are super influential and therefore the corresponding pixel groups, so users can easily identify the most significant patterns affecting the image's reality/fakeness. For example in Fig.~\ref{fig:teaser} D, it is clear that dimension 10 is one main factor leading to predicting a fake ``brown horse, side shot'' image as real, users can easily see this in the waterfall chart.

Moreover, the waterfall chart features a ``what-if'' button in its top-right corner. Activating this button presents the minimal changes needed across dimensions to alter the prediction outcome, offering a counterfactual analysis \textbf{(G4)}. This feature enables users to test their hypotheses, such as the impact of dimension 10, by observing the potential changes required to flip the prediction for confirmation.

\subsection{Dimension View}
To support counterfactual analysis and comparative analysis from a quantitative perspective \textbf{(G4, G5)}, we introduce the dimension view (Fig.\ref{fig:teaser} C). This view allows users to examine and compare the quantitative attributes across cells containing similar images, thereby deepening their understanding of the roles played by each dimension and the patterns associated with them at a subset level.

The dimension view incorporates two main components: the dimension value plot and the dimension contribution plot. The two plots were designed to help users assess quantitative correlations between visual patterns and the latent distributions across 16 dimensions as well as comprehend the contribution of individual dimensions within the distilled space.

\textbf{The dimension value plot} utilizes small multiples to present the dimension values of real and synthetic images' distilled representations across each of the 16 dimensions. Upon data initialization, it updates to show the global distribution of real versus synthetic images for each dimension within the distilled space. Dimensions are ordered based on the Kullback-Leibler divergence (KL divergence) between the distributions of real and fake images, with those showing the least divergence appearing first. This ordering emphasizes dimensions with closely overlapping real and synthetic distributions, indicating that pixel groups originating from these dimensions warrant closer examination. As illustrated in Fig.~\ref{fig:teaser} C, dimension 1 exemplifies a scenario where real and fake images share almost identical distributions, even within the specific cell of ``brown horse, side shot'' images, showcasing no discernible separation between correctly classified and misclassified images in that dimension. When a user selects a specific cell for analysis, the dimension value plots refresh to overlay the selected cell's distribution charts atop the global distribution. This enables users to instantly gauge how a cell's distribution deviates from the global norm.

The \textbf{dimension contribution plot} accessible via the ``show contribution plot'' button in the Dimension View's top-right corner, utilizes violin plots to represent the distribution of each dimension's contribution at both the global and subset levels. This is complemented by a boxplot that reveals key statistical metrics on hover. This visualization enables users to identify the primary dimensions that account for the differences between two groups of visually similar images with divergent prediction outcomes \textbf{(G5)}, guiding further explorations into the implications of these dimensions in the Pattern View.

Upon selecting a cell, the contribution plot incorporates parallel coordinates to display each image's distilled representation within the cell. Users can use the ``show correct'' and ``show incorrect'' buttons to filter the contribution distribution of correctly and incorrectly classified images within the cell, facilitating a comparison to pinpoint the critical dimensions contributing to prediction discrepancies between groups of images (as illustrated in Fig.~\ref{fig:violinplot}).

%% file: Sections/Usage_Scenario.tex
\section{Usage Scenario}
We showcase our system through two application scenarios, viewed from the perspective of a hypothetical user named \user, who employs \sys to analyze various synthetic patterns produced by different generative models. In the first scenario, \user examines synthetic patterns in horse images created by the proGAN model. In the second scenario, he investigates synthetic human faces generated by the Latent Diffusion Model (LDM).

\subsection{Analyzing patterns of horse images from proGAN}
\user initiates the analysis by selecting the proGAN model and the horse subject in the interface, then clicks ``load.'' Once the loading is complete, the interface presents a representation overview of images that are organized into local cells. Notably, the central region appears pale, suggestive of a decision boundary, with numerous orange cells surrounded by red arcs, indicative of deceptive synthetic patterns \textbf{(G5)}.

\user selects a prominent orange cell in the center, which refreshes the image view to display multiple synthetic images of brown horses, all captured from the side (Fig.~\ref{fig:teaser}). Many of these synthetic images are incorrectly classified as real, with some appearing convincingly realistic to him. Consequently, \user tags this cell as ``brown horse, side shot'' and proceeds to investigate adjacent cells, most of which contain still images of brown horses in side views \textbf{(G5)}. Moving through the cells, \user encounters one with a significantly high misclassification rate, primarily featuring the heads of black or brown horses, and tags it as ``brown/black horse head'' (Fig.~\ref{fig:cell_glyph}). Further exploration reveals a cell filled with misclassified images depicting riders on horses (Fig.~\ref{fig:violinplot}), which \user tags accordingly as ``rider on horses.''

A deeper investigation begins with the ``brown/black horse head'' cell (highlighted in Fig.~\ref{fig:violinplot} H1), where an adjacent cell also with ``brown/black horse head'' images exhibits markedly higher accuracy (seen in Fig.~\ref{fig:violinplot} H4). Employing the ``show concept'' feature, the user identifies key pixel groups in both cells, such as brownish fur patterns, sky patterns, and horse face patterns. With little discernible difference from the concept view, the user turns to the dimension contribution plot (referenced in Fig.~\ref{fig:violinplot}). A stark difference in dimension 2's contribution between the two cells becomes apparent (detailed in Fig.~\ref{fig:violinplot} H2 and H5). The pattern view indicates that pixel groups linked to dimension 2, which closely correlate with areas adjacent to or resembling horse eyes (noted in H3 and H6), significantly impact the classification \textbf{(G6)}. In misclassified images, these pixel groups exert a stronger push towards the ``real'' class (on average -0.07), contrasting with the weaker contributions in accurately classified images (on average 0).

Applying similar analytical methods, \user reviews the ``riders on horse'' cell, also with low sensitivity. A comparison with an adjacent cell, correctly identifying numerous ``rider on horse'' images, reveals a discrepancy in dimension 7 values, suggesting that realistic background patterns can sway the classification towards or away from being identified as synthetic \textbf{(G4, G6)}.

Zooming out to explore more intriguing cases, \user discovers a red-bordered blue cell amidst orange cells, indicating a significant number of real images misclassified as synthetic. Upon investigation, these images are predominantly real horse images from 3D games, with dimensions 3 and 16 being major contributors to the synthetic classification. Further analysis reveals that these dimensions are associated with the horse's body, suggesting that 3D graphics-generated horses lack a certain realism, hence their elevated fake scores.

\begin{figure}[htbp!]
    \centering
    \includegraphics[width=\linewidth]{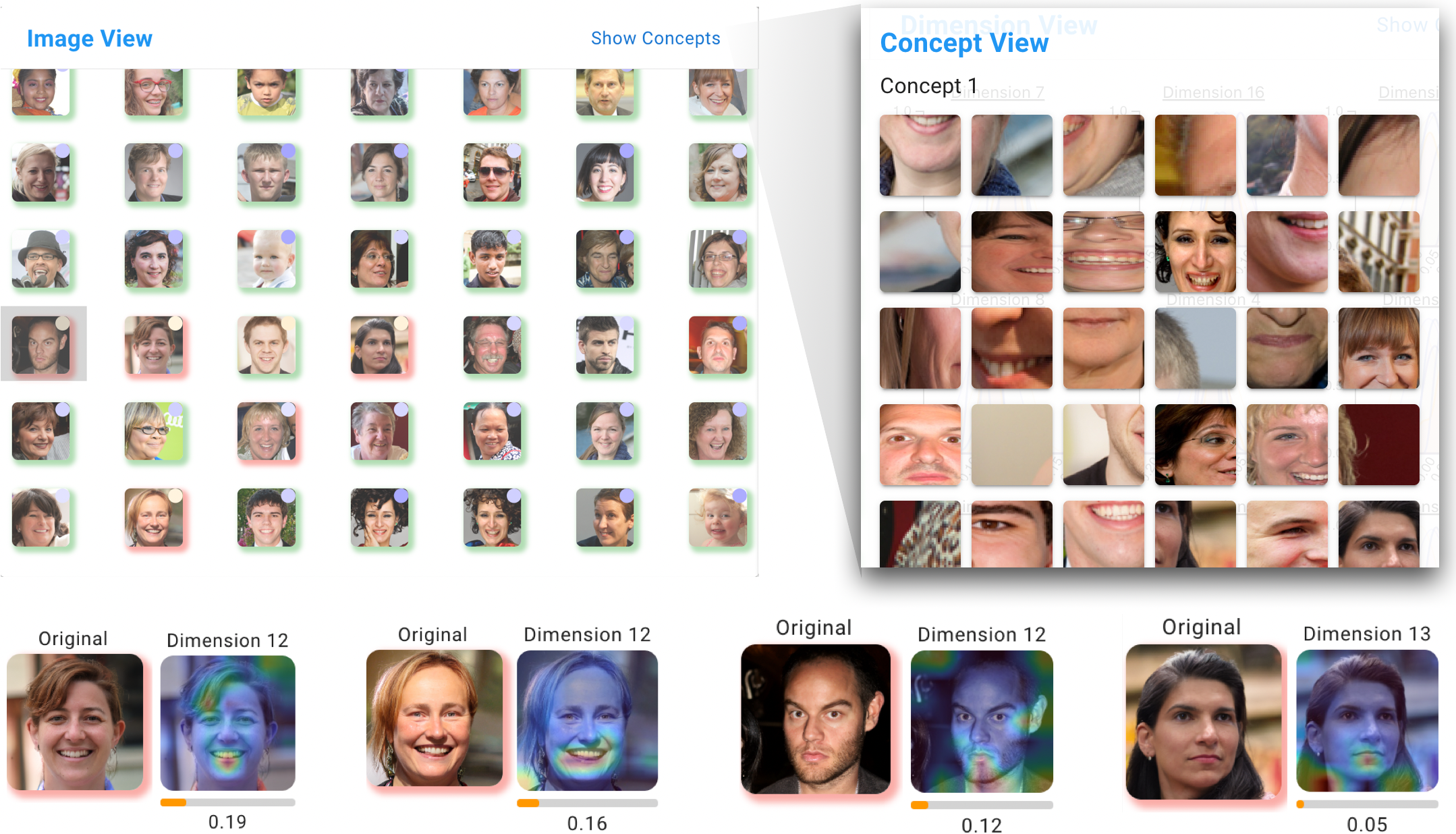}
    \caption{Leveraging \sys, the user identifies a cell containing misclassified images; upon examination, these are predominantly smiling faces. Through the concept view, it becomes evident that the mouth significantly influences the classification. However, a closer look at these misclassified images reveals that their smiles, though influencing classification were accurately detected by the model as their contribution are positive. This suggests the classifier was able to recognize 'fake smiles.' Further investigation shows that those images are misclassified due to high positive contribution from the forehead region}
    \label{fig:concept_smile}
\end{figure}

\subsection{Analyzing patterns of human face from LDM}

\user extends his exploration into synthetic human faces produced by the Latent Diffusion Model (LDM). Selecting LDM as the model and ``human face'' as the subject in the interface starts the analysis.

Similar to horse images, a pronounced decision boundary is identified in the Representation Overview, distinguishing real images from synthetic ones. The confusion matrix review reveals that, although the classifier effectively differentiates real from synthetic faces in general, it errs by classifying some synthetic faces as real and, more critically, mislabeling some real faces as synthetic. This means that even sophisticated classifiers are not infallible.

As before, the user tags several noteworthy cells for closer examination, including one particularly challenging cell with a low sensitivity rate filled with misclassified synthetic images featuring ``things on faces,'' and another populated with images of smiling individuals, labeled ``smiling faces.'' \textbf{(G5)}

\begin{figure}[htbp!]
    \centering
    \includegraphics[width=\linewidth]{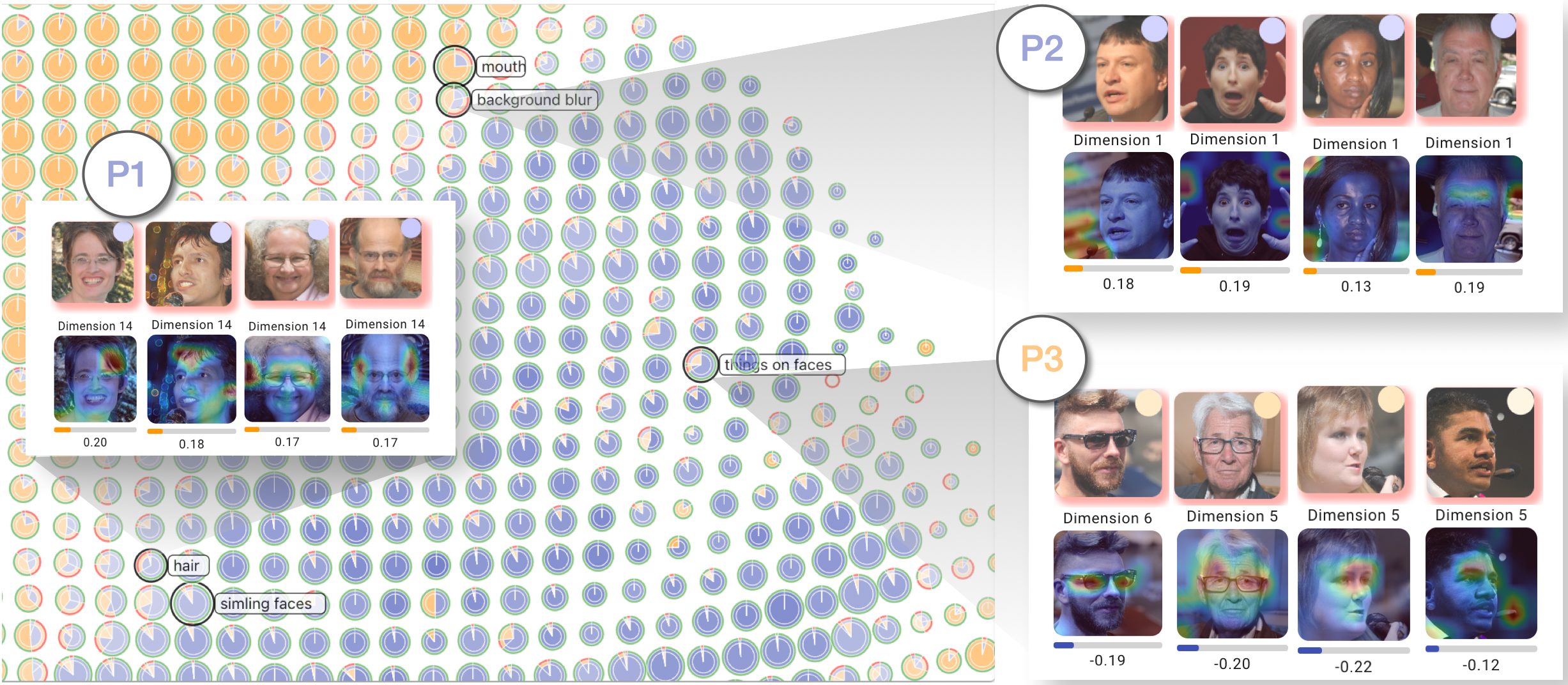}
    \caption{The user identifies common patterns in misclassified real faces, including hair (P1) and background blur (P2), as well as factors contributing to the misclassification of fake images, such as ``things on faces'' (P3).}
    \label{fig:hf_patterns}
\end{figure}

\textbf{``Things on faces'' cell}: This cell consists of images where human faces are associated with realistic items such as sunglasses, eyeglasses, microphones, and makeup. A detailed investigation reveals that these near-face objects significantly sway the classifier towards predicting these images as real (Fig.~\ref{fig:hf_patterns} P3) \textbf{(G5, G6)}.

\textbf{``Smiling Faces'' Cell}: This cell is filled with images of smiling faces. The concept view reveals that the mouth region appears to significantly influence the classifier's decision-making process. Surprisingly, the user discovers that the mouth region in these misclassified images contributes heavily towards a classification of fakeness. Subsequent scrutiny suggests that misclassifications often stem from the forehead region, unveiling another prevalent pattern of deception (Fig.~\ref{fig:concept_smile}) \textbf{(G4)}.

In addition to analyzing fake image patterns, \user also investigates real images that are flagged as synthetic \textbf{(G4)}. By examining cells with high false-negative rates (predominantly blue with red borders), the user identifies recurring misclassification patterns: real images are often misjudged based on hair features associated with dimension 14 (Fig.~\ref{fig:hf_patterns} P1) or background blur associated with dimension 1 (Fig.~\ref{fig:hf_patterns} P2) \textbf{(G5)}. This revelation points to the classifier's vulnerabilities, suggesting that to compromise an image's perceived authenticity, one might only need to introduce specific patterns to trigger a false positive classification.

%% file: Sections/Discussion_and_Future_Work.tex
\section{Discussion and Conclusion}

\sys represents a novel pipeline and interface for detecting AI-generated images and facilitating subsequent analysis and summarization of patterns inherent to such images. In particular, \sys helps address three main challenges faced by existing detection tools, namely a lack of (i) generalizability, (ii) interpretability, and (iii) accessibility. To overcome the challenge of limited generalizability, we propose the utilization of large, pre-trained models. These models have undergone extensive training on diverse datasets and have acquired robust representations of images. We leverage these capabilities to conduct classification based on their visual encodings. To address the issue of interpretability, we introduce a technique for assessing the influence of individual pixels, alongside a method for evaluating pixel groups. This approach aids in uncovering recurrent and deceptive patterns in AI-generated images. To enhance tool accessibility, we have organized the patterns identified by our backend into an interactive interface. This allows users to explore, analyze, and summarize the deceptive patterns of fake images interactively. Via a pair of usage scenarios, we demonstrate the utility of \sys for such workflows and highlight how it supports the design goals \textbf{(G1)}--\textbf{(G6)}.

To our knowledge, \sys represents a first-of-its-kind system for HITL analysis of GenAI content and models. During the design, development, and validation of \sys, we identified several aspects of our approach that can either generalize to future community efforts in these areas or suggest new avenues for future exploration:

\textbf{Interactive base model selection.} Our system currently employs the CLIP model as the foundational generic visual feature encoder to preprocess images. It then trains classifiers to distinguish between real and fake images, leveraging the knowledge gained to identify fake patterns. Given the release of increasingly powerful multi-modal pre-trained models on large-scale datasets, there exists a significant opportunity to support the interactive selection of base models for visual information encoding. This enhancement could further improve the downstream classifier's ability to identify more nuanced patterns.

\textbf{Analyzing other types of digital content.} In the era of advanced generative models, the production of synthetic content extends beyond images to include text, music, and videos. While \sys presently focuses on image analysis, the underlying framework—utilizing a base model as a generic encoder and training a specialized classifier to detect fake patterns—can be generalized to other modalities. Our token relevance computation method, compatible with all types of transformer models, remains applicable as long as the input data is encoded using a transformer-based model.

\textbf{Developing tools to help developers.} Though \sys is designed to aid professionals in understanding and detecting fake AI-generated content, the visualization designs we develop are generalizable to contribute to the development and usage of more sophisticated detection tools. As the landscape of generative models evolves, with increasingly complex models being introduced, the need for advanced detection mechanisms becomes more apparent. For example, the representation overview in our tool can assist developers in identifying the strengths and weaknesses of these emerging detection systems.

\textbf{Facilitating prompt engineering.} Beyond its primary role in combating the adverse effects of GenAI, our tool also offers value to developers utilizing GenAI in their projects, such as photographers, graphic designers, and other creatives using AI to enhance their work. With the dominance of text-to-image generative models, tools like \sys can be extended with prompting-specific functionalities to help identify optimal prompts for generating realistic images.